\documentclass[review]{elsarticle}
\usepackage{lineno}
\modulolinenumbers[5]
\usepackage{xcolor}
\usepackage[colorlinks = true,
            linkcolor = red,
            urlcolor  = red,
            citecolor = green,
            anchorcolor = blue]{hyperref}
\hypersetup{
 colorlinks=true,
 citecolor=Violet,
 linkcolor=Red,
 urlcolor=Blue}
\journal{NeuroImage}
\usepackage{comment}
\graphicspath{{Figures/}}
\usepackage{graphicx}
\usepackage{tabulary}
\usepackage{booktabs}
\usepackage{subcaption}
\usepackage{amssymb}
\usepackage{amsmath}
\usepackage[colorinlistoftodos]{todonotes}
\usepackage{xcolor}
\usepackage{mdframed}
\usepackage[inline]{enumitem}
\biboptions{authoryear}
\begin{document}

\begin{frontmatter}

\title{Correcting differences in multi-site neuroimaging data using Generative Adversarial Networks}
%\tnotetext[mytitlenote]{Fully documented templates are available in the elsarticle package on \href{http://www.ctan.org/tex-archive/macros/latex/contrib/elsarticle}{CTAN}.}

%% Group authors per affiliation:
\author[address1]{Harrison Nguyen}
% * <richard.morris@sydney.edu.au> 2018-01-09T03:45:02.328Z:
% 
% > Harrison T. Nguyen
% You are listed as first author because you did most of the work, and the remaining authors are listed in order of contribution - except the final author (Fabio) who takes the senior author position and will be ultimately responsible for the work
% 
% ^.
\author[address2,address3]{Richard W. Morris}
\author[address2,address4]{Anthony W. Harris}
\author[address2,address4]{Mayuresh S. Korgoankar}
\author[address1,address3]{Fabio Ramos\corref{mycorrespondingauthor}}
% * <richard.morris@sydney.edu.au> 2018-01-09T03:42:09.999Z:
% 
% > \corref{mycorrespondingauthor}
% Fabio is corresponding author because it is his lab you did the work in and he is ultimately responsible for the accuracy. His contact details are also most likely to be correct as well (since yours will change when you leave the lab), so that others can follow-up the work once it is published.
% 
% ^.
\cortext[mycorrespondingauthor]{Corresponding author}
\ead{fabio.ramos@sydney.edu.au}
\ead[url]{https://sydney.edu.au/engineering/people/fabio.ramos.php}

\address[address1]{School of Information Technologies, University of Sydney, Sydney, Australia}
\address[address2]{School of Medicine, University of Sydney, Sydney, Australia}
\address[address3]{Centre for Translational Datascience, University of Sydney, Sydney, Australia}
\address[address4]{Brain Dynamics Centre, Westmead Millennium Institute, Sydney, Australia}
% \listoftodos
\begin{abstract}
Magnetic Resonance Imaging (MRI) of the brain has been used to investigate a wide range of neurological disorders, but data acquisition can be expensive, time-consuming, and inconvenient. Multi-site studies present a valuable opportunity to advance research by pooling data in order to increase sensitivity and statistical power. However images derived from MRI are susceptible to both obvious and non-obvious differences between sites which can introduce bias and subject variance, and so reduce statistical power. To rectify these differences, we propose a data driven approach using a deep learning architecture known as generative adversarial networks (GANs). GANs learn to estimate two distributions, and can then be used to transform examples from one distribution into the other distribution. Here we transform T1-weighted brain images collected from two different sites into MR images from the same site. We evaluate whether our model can reduce site-specific differences without loss of information related to gender (male or female) or clinical diagnosis (schizophrenia or healthy). When trained appropriately, our model is able to normalise imaging sets to a common scanner set with less information loss compared to current approaches. An important advantage is our method can be treated as a `black box'  that does not require any knowledge of the sources of bias but only needs at least two distinct imaging sets. 
% * <richard.morris@sydney.edu.au> 2018-01-10T00:30:15.567Z:
% 
% Our method is general to MRI (structural, functional, DTI etc), so let's not focus on sMRI in our exposition - sMRI is just our proof-of-concept (admittedly a much easier example than the others)
% 
% ^.
\end{abstract}


\end{frontmatter}

%\linenumbers

\section{Introduction}
One of the biggest challenges in the translation of neuroimaging findings into clinical practice is the need to validate models across large independent samples and across data obtained from different MRI scanners and sites. Combining multiple samples increases the overall sample size, overcoming a limitation common to many neuroimaging studies. However it also introduces heterogeneity into the sample from differences in scanner manufacturer, MRI protocol, variation in site thermal and power stability, as well as site differences in gradient linearity, centering and eddy currents. Therefore, images from different sites have the potential to introduce bias that can either mimic or obscure true changes or even worse, produce results that could be driven by the artifactual site differences. This can make the interpretation, reliability and reproducibility of findings difficult. Despite these issues, pooling data provides the opportunity to address a major source of concern regarding the low statistical power of published studies, especially when larger studies are not feasible due to financial constraints or recruitment is difficult because a particular disorder is rare at a specific geographical location \citep{poldrack2014making}.

Given the considerable incentives to pool data, there is a relative paucity of methods available to correct for site-specific differences in MR images. The majority of approaches are usually applied during data acquisition, for instance, using a common phantom across sites to calibrate and reduce differences in field homogeneities. However, these \textit{a priori} methods require careful planning and are not applicable to data sets that have already been collected or other \textit{post hoc} forms of data pooling. Site differences can also be addressed in a \textit{post hoc} fashion by treating the site as a covariate in the analysis for evaluation of confounding effects. However, the interaction between the usually unknown site-specific effects and the true brain effects on the MRI signal seem to be highly complex and nonlinear such that the inclusion of the covariate can also introduce bias \citep{rao2017predictive}.

Recent advances in computer vision due to the application of artificial neural networks suggests there may be a novel \textit{post hoc} solution  to remove non-linear bias in MR images. For example, superior performance in non-linear, multivariate pattern classification problems such as Alzheimer’s disease classification, brain lesion segmentation, skull stripping and brain age prediction have been achieved using deep learning networks \citep{payan2015predicting, sarraf2016deepad, kamnitsas2017efficient, kleesiek2016deep, cole2017predicting}. Deep learning provides some unique advantages for high-dimensional data such as MRI data, since it does not require extensive feature engineering. Furthermore, deep learning has produced important advances in generative modeling. Generative modeling involves learning to estimate a given distribution in order to produce examples from that distribution. For example, after being trained on a set of images, the model is able to generate a new, `unseen’ sample from the training set. Generative modeling is considered a much more difficult task than pattern classification, as the output of these models are typically high dimensional and a single input may correspond to many correct answers (e.g. there are many ways of producing an image of a cat). 

\begin{figure}[htp]
\begin{center}
 \includegraphics[width = 0.9\textwidth]{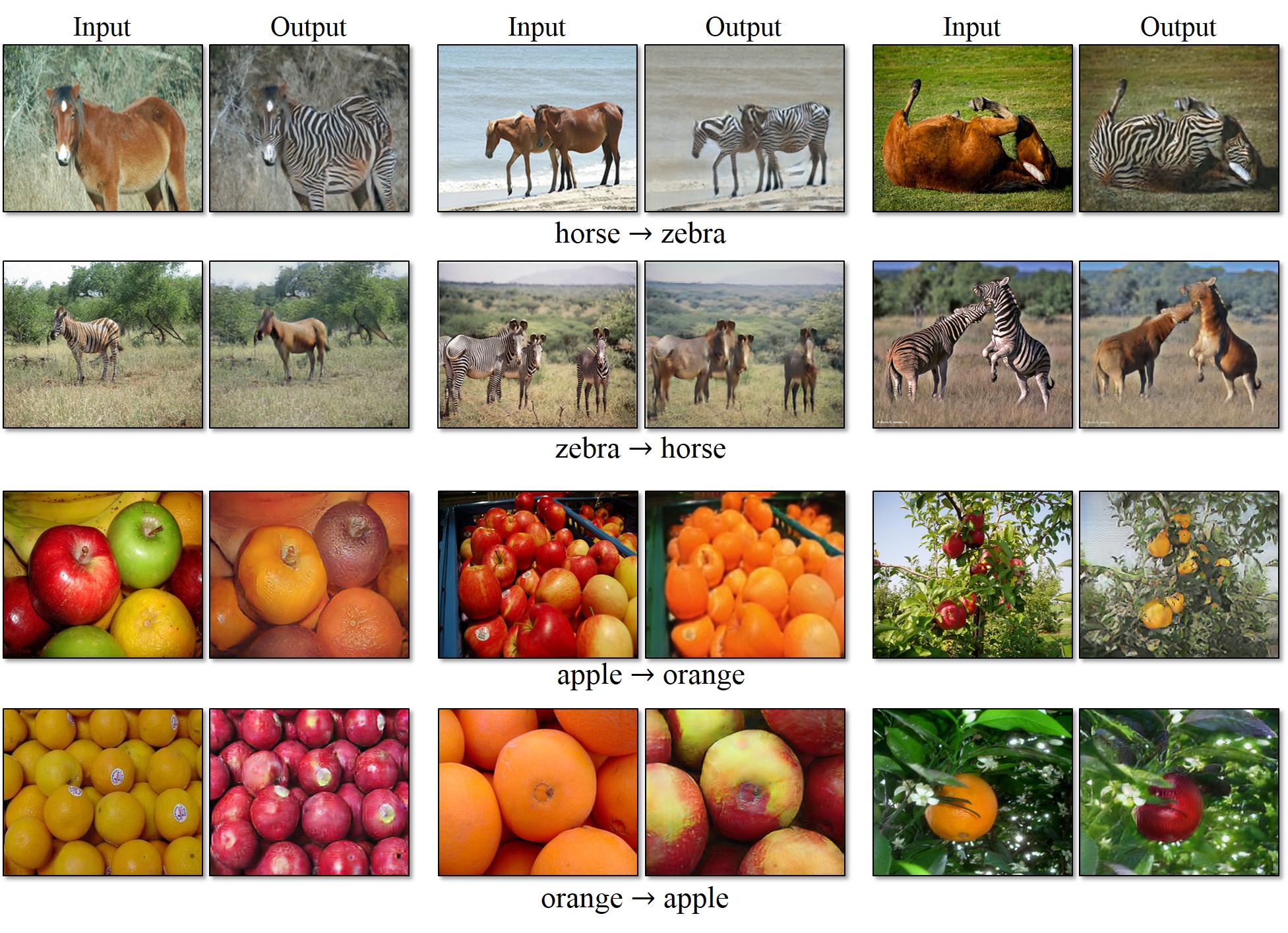}
    \end{center}
  \caption{Examples of images produced from CycleGAN. Reproduced from  \cite{zhu2017unpaired} without adaption, under CC-BY 4.0 }
  \label{fig:cycle_gan}
\end{figure}

One class of generative models, known as generative adversarial networks (GANs), have recently achieved considerable success in a variety of image problems, from image generation \citep{radford2015unsupervised}, super resolution generation \citep{ledig2016photo}, text2image \citep{reed2016generative} and image-to-image translation \citep{isola2016image} (See Figure \ref{fig:cycle_gan} for examples). GANs succeed through the idea of adversarial training, where the model’s training process can be described as a game between two players. One player is called the generator where it attempts to create samples from the same distribution as the observed data. The other player is the discriminator where its function is to examine the fake samples from the generator and real samples from the observed data and to classify the generated and observed samples as either real or fake. Over time, the discriminator is trained with supervision to better distinguish real and fake samples. However at the same time, the generator will improve its synthesis of fake samples in order to fool the discriminator, which in turn will make the job of the discriminator more difficult. Eventually the solution of this game is a Nash equilibrium, where the generator is unable to improve its generation of fake samples and the discriminator is unable to better classify real and fake samples \citep{goodfellow2016nips}. See Box 1 for further details.

Here, we propose an algorithm that uses GANs to transform a set of images from a given MRI site into images with characteristics of a different MRI site. Its purpose is to correct for differences in site artifacts without the need for \textit{a priori} calibration using phantoms or significant coordination of acquisition parameters. This algorithm can be treated as a ’black box’ without knowledge of the artifacts present in the dataset and can be applied \textit{post hoc} after acquisition to two or more unpaired sets of imaging data. Importantly, as we demonstrate, the correction occurs without any apparent loss of information related to gender or clinical diagnosis.

\section{Material and methods}
This research was conducted under approval from the University of Sydney Human Research Ethics Committee, HREC 2014/557.
\subsection{Participants}
Structural (T1-weighted) MR brain images were obtained (N = 313) from pre-existing MRI studies conducted at two different sites (site A and site B). The cohort from each site contained two diagnostic groups (schizophrenia and healthy adults), however these groups were not evenly distributed over sites (see Table \ref{tab:cohort}). All clinical cases met DSM-IV criteria for their disorder with no other Axis I disorders, on the basis of either the Mini-International Neuropsychiatric Interview \citep{hergueta1998mini} or the Structured Clinical Interview for DSM-IV Axis I and II Disorders \citep{first2002structured}. Participants were aged 18-65 years and spoke fluent English. Exclusion criteria included the presence of an organic brain disorder, brain injury with post-traumatic amnesia, mental retardation (WAIS-III IQ score less than 80), movement disorders and recent (within 6 months) substance dependence or electroconvulsive therapy. Healthy adults were also screened for the absence of personal or family history of any DSM-IV Axis I disorder. 

\begin{table}[htbp]
  \centering
  \caption{Subject and gender distribution across sites (m:male, f:female)}
    \begin{tabular}{l|lll}
    \toprule
     & \textbf{Site A} & \textbf{Site B} &\textbf{Total} \\
    \midrule
    Control \\ 
    \hspace{1em}n &41 &101 &142 \\
    \hspace{1em}age $\pm$ SD    &29.7$\pm$13.1 &31.2$\pm$8.7 & 31.2$\pm$10.1	  \\
    \hspace{1em}m/f   & 23/18 & 52/49 & 75/67\\
     Schizophrenia\\
     \hspace{1em}n &17 &154 &171 \\
     \hspace{1em}age $\pm$ SD    &44.8$\pm$11.1 &38.0$\pm$9.5	 & 38.7$\pm$9.8  \\
     \hspace{1em}m/f   & 7/10 & 57/97 & 64/107\\
     %Bipolar \\
     %\hspace{1em}age $\pm$ SD    &  24.5$\pm$5.4&43.3$\pm$11.4& 34.5$\pm$13.1 	  \\
    %\hspace{1em}m/f   & 12/9 & 12/12 & 24/21\\
     Total \\
     \hspace{1em}n &58 &255 &313 \\
     \hspace{1em}age $\pm$ SD    & 34.1$\pm$14.1&	35.3$\pm$9.7 & 35.1$\pm$10.7\\
    \hspace{1em}m/f   & 30/28 & 109/158 & 139/174\\
    \bottomrule
    \end{tabular}%
  \label{tab:cohort}%
\end{table}

\subsection{MR Scanner, image data and preprocessing}
Data were collected from two different MRI sites: Site A hosted a Phillips Achieva 3T with a 8-channel headcoil and receiver (NeuRA, Randwick NSW, Australia); and Site B hosted a GE Discovery MR750 3T with a 8-channel headcoil and receiver (Brain and Mind Centre, Camperdown NSW, Australia). T1-weighted image volumes were acquired using a standard but scanner-specific MPRAGE acquisition sequence. T1 images from Site A were acquired with a 3D Fast Spoiled Gradient Recall Echo (FSPGR) sequence with SENSE acceleration; 8.3-ms TR, 3.2-ms TE; and 11 degree flip angle, and comprised of 180 sagittal 1-mm slices in a 256 x 256 matrix (1 mm isotropic voxel dimensions). Images from Site B were acquired with a 3D Turbo Field Echo sequence (TFE) with ASSET acceleration; 7.192-ms TR, 2.732-ms TE; and 12 degree flip angle, and comprised of 176 sagittal 1-mm slices in a 256 x 256 matrix (1 mm isotropic voxel dimensions). 

%  according to a widely adopted procedure described in Ashburner (2010, "optimized VBM"). 
Image preprocessing was designed to remove as much of the site differences as possible given standard tools available, before applying the novel GAN method described in the next section. All preprocessing occurred in SPM12 (http://www.fil.ion.ucl.ac.uk/spm), running under Matlab 8.4 (Math Works, Natick, MA, USA). After checking for scanner artifacts and gross anatomical abnormalities for each image, we reoriented the original images along the anterior-posterior commissure (AC-PC) line and set the AC as the origin of the spatial coordinates to assist the normalization algorithm. The unified segmentation procedure in SPM12 was used to segment all the images into mean corrected gray matter (GM), white matter (WM) and cerebrospinal fluid (CSF) space, i.e. maps of probability values representing the probability of a voxel containing a specific tissue type. Mean correction was applied to remove site differences in the bias field. A fast diffeomorphic image registration algorithm \citep{ashburner2007fast} was used to warp the GM partitions into a new study-specific reference space representing an average of all 313 subjects included in the analysis. As an initial step, a set of study-specific templates and the corresponding deformation fields, required to warp the data from each subject to the new reference space, were created using the GM partitions \citep{ashburner2009computing}. Each subject-specific deformation field was used to warp the corresponding GM partition into the new reference space with the aim of maximizing accuracy and sensitivity \citep{yassa2009quantitative}; the warped GM partitions were affine transformed into the MNI space and an additional `modulation' step was used to scale the GM probability values by the Jacobian determinants of the deformations in order to ensure that the total amount of gray matter in each voxel was conserved after the registration \citep{ashburner2000voxel,good2001cerebral,mechelli2005structural}. After this preprocessing, we obtained bias-field corrected, modulated, normalized gray matter density maps, from which we extracted the middle five 2D sagittal slices to be used to train the GAN model described below.

\subsection{Generative Adversarial Networks}
Rather than removing any remaining scanner artifacts and biases from the images, we seek to transform one set of images from a site to images that come from the distribution of images from the other site, while still preserving the important features of the original images. 

\textbf{Notation}: In the following, we have defined capital bold font, $\mathbf{X}$, as a matrix or a set of images and lower case bold font, $\mathbf{x}$, as a vector or one example image. $G_\mathbf{\theta}$, $D_\mathbf{\phi}$ denotes a mapping function parameterised by $\mathbf{\theta}$ and $\mathbf{\phi}$, respectively. $P(\mathbf{X})$ indicates the probability distribution for the imaging set $\mathbf{X}$, and $\hat{P}(\mathbf{X})$ is an estimate of that probability distribution.

The problem at hand can be described as image-to-image translation in the computer vision literature where the goal is to learn a mapping function between a set of MRI images from domain $\mathbf{X}$ and another set of images from domain $\mathbf{Y}$; learn $G: \mathbf{X}\rightarrow \mathbf{Y}$ such that $G(\mathbf{x})$ for each $\mathbf{x} \in \mathbf{X}$ is indistinguishable from the set of images from domain $\mathbf{Y}$. 

The CycleGAN \citep{zhu2017unpaired} and DiscoGAN \citep{kim2017learning} have been developed to learn cross domain relationships between sets of natural objects such as from horses to zebras, edges to photos and Monet artworks to realistic photos. The advantage of these models is that they do not require paired sets of training samples, $\{\mathbf{x}_i,\mathbf{y}_i\}^N_{i=1}$, which is often difficult to obtain for neuro-imaging data, and instead only require unpaired imaging data consisting of a source set $\{\mathbf{x}_i\}^N_{i=1}\in \mathbf{X}$ and target set $\{\mathbf{y}_j\}^M_{j=1} \in \mathbf{Y}$ , without any $\mathbf{x}_i$'s necessarily corresponding to any $\mathbf{y}_j$'s. These models attempt to transform the underlying distribution of $P(\mathbf{X})$ to an estimate of $P(\mathbf{Y})$, $\hat{P}(\mathbf{Y})$, through $G$ while still preserving the important features of the original sample, $\mathbf{x}_i$, but also merging these with the particular characteristics of $P(\mathbf{Y})$.

To learn this mapping function, an adversarial training regime was utilised using the GAN formulation. The generator, $G_\mathbf{\theta}$, represented as a convolutional neural network defined by parameters $\mathbf{\theta}$, takes as input, images from $\mathbf{X}$ and transforms these images, $G_\mathbf{\theta}(\mathbf{x})$, as if they were sampled from $P(\mathbf{Y})$. The discriminator, $D_\mathbf{\phi}$ on the other hand, is a supervised classifier represented as a convolutional neural network. The discriminator observes two inputs, the observed images from $\mathbf{Y}$ and generated samples $G_\mathbf{\theta}(\mathbf{x})$. The goal of the discriminator is to output a probability that its inputs are either real or fake, with the true labels being observed samples as real and generated samples as fake. The discriminator attempts to learn that its output from samples of $\mathbf{Y}$,  $D_\mathbf{\phi}(\mathbf{y})$, are given to be given values near 1 and inputs from the generator, $D_\mathbf{\phi}(G_\mathbf{\theta}(\mathbf{x}))$, to be values close to 0. However, at the same time, the generator will attempt to make the quantity, $D_\mathbf{\phi}(G_\mathbf{\theta}(\mathbf{x}))$ to approach 1. At equilibrium, $D_\mathbf{\phi}(\mathbf{y})=\frac{1}{2}$ for all $\mathbf{y}$ and $G_\mathbf{\theta}(\mathbf{x})$ which means that the discriminator is unable to distinguish between real and generated samples.
 
The generator and discriminator face two competing objectives during training; the discriminator attempts to push $D_\mathbf{\phi}(G_\mathbf{\theta}(\mathbf{x}))$ to 0 and whilst on the other hand, the generators strives to fool the discriminator and make $D_\mathbf{\phi}(G_\mathbf{\theta}(\mathbf{x}))$ equal to 1. 

More specifically, the Least Squares GAN (LSGAN) \citep{mao2016least} is used to train the discriminator and generator, where the discriminator's objective function is
\begin{equation}
	\label{eqn:discrim_loss}
    \min_\mathbf{\phi} \frac{1}{2}E_{\mathbf{y}\sim p(\mathbf{Y})}[(D_\mathbf{\phi} (\mathbf{y}) - 1)^2] + \frac{1}{2}E_{\mathbf{x}\sim p(\mathbf{X})}[(D_\mathbf{\phi} (G_\mathbf{\theta}(\mathbf{x})))^2],
\end{equation}
and the generator competes against the discriminator by having the objective function 
\begin{equation}
	\label{eqn:gen_loss}
    \min_\mathbf{\theta} \frac{1}{2}E_{\mathbf{x}\sim p(\mathbf{X})}[(D_\mathbf{\phi} (G_\mathbf{\theta}(\mathbf{x}))-1)^2].
\end{equation}
Equation \ref{eqn:discrim_loss} and \ref{eqn:gen_loss} is typically optimised using stochastic gradient decent where $\mathbf{\phi}$ is updated keeping the generator's parameters fixed for one or more iterations and vice versa. Details about the training parameters are described in Section \ref{implementation}.  Equation \ref{eqn:discrim_loss} is optimised in a supervised manner where the ground truth labels, real or fake, are provided to the discriminator through the inputs $\mathbf{y}$ and $G_\mathbf{\theta}(\mathbf{x})$ respectively. Mao et al. demonstrated that minimising the objective function of LSGAN yields minimising the Pearson $\chi^2$ divergence between $\mathbf{Y}$ and $G_\mathbf{\theta}(\mathbf{x})$ \citep{mao2016least}.

\begin{comment}
The discriminator learns a decision boundary between real and fake samples and although some fake samples might be correctly classified, Equation \ref{eqn:discrim_loss} penalises samples further away from the decision boundary. Since $\mathbf{\phi}$ is fixed when updating the generator, the the decision boundary learned by the discriminator is also kept fixed. The dependence between the generator on the discriminator for learning (Equation \ref{eqn:gen_loss}) encourages the generator to produce samples closer to the decision boundary. On the next iteration, this causes the discriminator to update its decision boundary closer to the real data's manifold. The process repeats, making the decision boundary to pass through the real data manifold and the generated samples closer to the manifold of real data.
\end{comment}

Equation \ref{eqn:gen_loss}, in contrast to the learning objective of the discriminator, shows the generator does not have the same level of supervision as the discriminator. While although they have competing objectives, the generator improves its generation of samples, not because of the directive by a supervisor but rather, by the information provided by the discriminator. It is through the cooperation between the generator and discriminator that the generator learns the mapping function in an unsupervised manner. This enables the ability to learn the transformation that is data driven and without any a-priori knowledge of the processes that generated the two image sets.

The GAN objectives is not limited to Equations \ref{eqn:discrim_loss} and \ref{eqn:gen_loss}. Other adversarial formulations have been developed in order to minimise other divergence measures between the observed distribution and generated distribution such as the $f$-divergence \citep{nowozin2016f}, Jensen-Shannon divergence \citep{goodfellow2016nips} or other distance metrics to have different geometric interpretations such as, and not limited to, Earth Mover distance \citep{arjovsky2017wasserstein} and Integral Probability Metrics \citep{mroueh2017mcgan}. 
Results based on the $f$-divergence, Jensen-Shannon divergence and Earth Mover distance were also included in experiments but produced similar results to the LSGAN. They have not been included for the sake of brevity.

\subsubsection{Cycle loss}
However, the transformation $G:\mathbf{X}\rightarrow \mathbf{Y}$ is ill-posed as there are infinitely many mappings, $G(\mathbf{x})$, that could induce the estimated distribution $\hat{P}({\mathbf{Y}})$. This means that each $\mathbf{x}$ and output $G(\mathbf{x})$ do not necessarily have to have any meaningful relationship.  For example, a possible outcome is that $G_\mathbf{\theta}$ learns to transform all $\mathbf{x} \in \mathbf{X}$ , to only one particular example of $\mathbf{Y}$. This outcome is known as mode collapse where the generator learns to map several different input values to the same output point that fools the discriminator and the model is unable to make any progress in training.

To prevent this issue from occurring, the model is required to be constrained to a one-to-one correspondence (bijective mapping) by introducing the idea of a \textit{cycle} loss \citep{zhu2017unpaired}. If we have a mapping $G: \mathbf{X}\rightarrow \mathbf{Y}$ and another mapping $F: \mathbf{Y}\rightarrow \mathbf{X}$ then $G$ and $F$ should be inverses of each other. To ensure this, the generators $G$ and  $F$ are both trained simultaneously with their own adversarial loss and own parameters, $\mathbf{\theta_1}$ and $\mathbf{\theta_2}$ respectively but also adding a loss that encourages $F_\mathbf{\theta_2}(G_\mathbf{\theta_1}(\mathbf{x})) \approx \mathbf{x}$ and $G_\mathbf{\theta_1}(F_\mathbf{\theta_2}(\mathbf{y})) \approx \mathbf{y}$. The generators $G_\mathbf{\theta_1}$ and $F_\mathbf{\theta_2}$ are able to reconstruct the original set of images. Any distance metric function (L$_1$, Huber loss, cosine) could be used but in particular, the L$_2$ norm was used, 
\begin{equation}
    \label{eqn: reconstuction_loss}
    L_{cycle}(G, F) =E_{\mathbf{x}\sim p(\mathbf{X})} [\| F_\mathbf{\theta_2}(G_\mathbf{\theta_1}(\mathbf{x})) - \mathbf{x}\|_2] +E_{\mathbf{y}\sim p(\mathbf{Y})} [\| G_\mathbf{\theta_1}(F_\mathbf{\theta_2}(\mathbf{y})) - \mathbf{y}\|_2].
\end{equation}
\begin{figure}[htp]
\begin{center}
 \includegraphics[width = 0.7\textwidth]{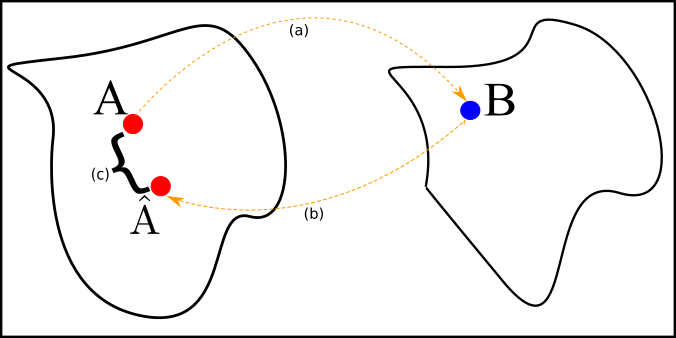}
    \end{center}
  \caption{\textbf{(a)}: Image A is mapped into the manifold of scanner set B through a a convolutional neural network (generator). \textbf{(b)}: This image is then transformed back to the original manifold to reconstruct the original image using a different CNN. \textbf{(c)}: The original and reconstructed image is compared using some distance metric (e.g. L$_1$ or L$_2$-norm).}
  \label{fig:pca_tsne}
\end{figure}
\subsubsection{Full objective}
The model contains two pairs of GANs, with each generator learning the respective mapping functions $G: \mathbf{X}\rightarrow \mathbf{Y}$ and $F: \mathbf{Y}\rightarrow \mathbf{X}$. Each generator will have their respective discriminators, $D_\mathbf{\phi_1}$ and $D_\mathbf{\phi_2}$, where $D_\mathbf{\phi_1}$ will discriminate between $\mathbf{x} \in \mathbf{X}$ and samples from $F_\mathbf{\theta_2}$ and conversely, $D_\mathbf{\phi_2}$ will distinguish between $\mathbf{y} \in \mathbf{Y}$ and the output of $G_\mathbf{\theta_1}$. The objective function of $G_\mathbf{\theta_1}$ and $D_\mathbf{\phi_2}$ is given respectively as
\begin{equation}
\label{eqn:full_eqn}
   \min_\mathbf{\theta_1} E_{\mathbf{x}\sim p(\mathbf{X})}[(D_\mathbf{\phi_2} (G_\mathbf{\theta_1}(\mathbf{x}))-1)^2] + \lambda L_{cycle}(G_\mathbf{\theta_1}, F_\mathbf{\theta_2}) 
\end{equation}
\begin{equation}
    \min_\mathbf{\phi_2} E_{\mathbf{y}\sim p(\mathbf{Y})}[(D_\mathbf{\phi_2} (\mathbf{y}) - 1)^2] + E_{\mathbf{x}\sim p(\mathbf{X})}[(D_\mathbf{\phi_2} (G_\mathbf{\theta_1}(\mathbf{x})))^2]
\end{equation}
where $\lambda$ is a constant that controls the relative importance between the adversarial loss and reconstruction loss.
The objective function for $F_\mathbf{\theta_2}$ and $D_\mathbf{\phi_1}$ are similarly defined.

\begin{comment}
\begin{figure}[htp]
\begin{center}
 \includegraphics[width = 0.9\textwidth]{cycle_set.png}
    \end{center}
  \caption{Generator A takes an image scan from scanner B and transforms that image into another, $\textbf{B}_\textbf{A}$, such that Discriminator A cannot distinguish it from an image from scanner A. Generator B then takes $\textbf{B}_\textbf{A}$ in order to reconstruct the original image. This process is mirrored for image scans from scanner A.}
  \label{fig:cycle_diagram}
\end{figure}
\end{comment}

\subsection{Implementation} \label{implementation}
The generators and discriminators are fully convolutional neural networks. The discriminators are composed of six convolutional layers to create a receptive field of 30$\times$30 patches that aims to classify whether 30$\times$30 overlapping image patches are either real of fake. The transformations of the input consists of a succession of spatial 2D convolutions, a transformation that keeps the input distribution of each hidden layer similar during training by normalising a training batch (batch normalisation) and a voxel-wise non-linear transformation (also known as an activation function) of the results of the convolutions. 

During training, the input distribution of each hidden layer may change after several iterations, known as internal covariate shift, due to the complicated non-linearities of the incoming neurons. The current hidden layers will have to continually adapt to these changes in the input distribution hence could slow down convergence. Batch normalisations attempts to rectify this by normalising the inputs to each hidden layer so that their distribution during training remains fairly constant \citep{DBLP:journals/corr/IoffeS15} which improves convergence of training. In regards to the choice of activation function, the \textit{leakyReLU} activation function was used as it was found to have the best qualitative performance except in the last layer of the discriminators where no activation function was used. 

The generators contain two convolutional down sampling layers, reducing the dimensionality of the image by a factor of four, followed by six residual blocks to create new features of the data then another two convolutional upsampling layers to restore the image back to the original input dimensions. The residual blocks is composed of two convolutional layers that includes a `skip' connection, where the input to these layers are added to the output of the convolution layers. The residual blocks are critical to the generator as some portions of the image may not necessarily require any transformations. Therefore, including these residual layers will give the option of the network to skip convolutional layers and not undergo any change. Much like the discriminator, each convolutional layer is followed by batch normalisation and then a \textit{leakyReLU} activation function. However, the last layer of the generator uses a \textit{tanh} function that scales the output from -1 to 1, producing a new grey matter voxel map. More specific details about the architecture used is found in Table \ref{tab:network_arch}.

\begin{table}[htbp]
  \centering
  \begin{subtable}{\linewidth}
  \caption{Architecture of Generator}
    \begin{tabular}{c|ccccc}
    \toprule
    \textbf{Layer} & \textbf{Layer Type} & \textbf{No. of Filters} & \textbf{Stride}&  \textbf{Batch Norm} &\textbf{Activation Function} \\
    \midrule
    1    & Convolution	 & 32& 2 & No& LeakyReLU \\
    2    & Convolution & 64 & 2  & Yes& LeakyReLU \\
    3    & Convolution & 128 & 2 &  Yes &LeakyReLU  \\
    4-6    & Residual Block & 128 & 1 & Yes & LeakyReLU \\
    7    & Convolution Transpose & 64 & 2& Yes  & LeakyReLU\\
    8    & Convolution Transpose & 32 & 2 & Yes & LeakyReLU\\
    9    & Convolution & 1 & 1  & No & Tanh\\
    \bottomrule
    \end{tabular}%
  \label{tab:generator_arichitecture}%
  \end{subtable}
    \begin{subtable}{\linewidth}
  \caption{Architecture of Discriminator}
    \begin{tabular}{c|ccccc}
    \toprule
    \textbf{Layer} & \textbf{Layer Type} & \textbf{No. of Filters} & \textbf{Stride}&  \textbf{Batch Norm} &\textbf{Activation Function} \\
    \midrule
    1    & Convolution	 & 32& 2 & No& LeakyReLU \\
    2    & Convolution & 64 & 2  & Yes& LeakyReLU \\
    3    & Convolution & 128 & 2 &  Yes &LeakyReLU  \\
    4    & Convolution  & 128 & 1& Yes  & LeakyReLU\\
    5    & Convolution & 128 & 1 & Yes & LeakyReLU\\
    6    & Convolution & 1 & 1  & No & None\\
    \bottomrule
    \end{tabular}%
  \label{tab:discriminator_arichitecture}%
  \end{subtable}%
  \caption{Architecture of Generative Neural Network}
  \label{tab:network_arch}%
\end{table}%

During training, mini-batches consisting of eight sagittal slices were constructed from each scanner set. The filters of the CNN were intialised as described by Glorot and Bengio \citep{glorot2010understanding}. The network was trained using Adam optimisation \citep{kingma2014adam} with a starting learning rate of 2e-4 for the generators and discriminators. The generators and discriminators were trained concurrently; every one gradient step of the generator was taken with the discriminator parameters fixed followed by a gradient step of the discriminator, keeping the generator parameters fixed. Training was stopped when the cycle loss (Equation \ref{eqn: reconstuction_loss}) failed to stop decreasing. It was found empirically that the hyperparamter, $\lambda$, in Equation \ref{eqn:full_eqn} was set to $\lambda = 0.2$, balancing between faster convergence and qualitative results.

\subsection{Postprocessing}
For better classification results as outlined in Section \ref{evaluation_methods}, Principal Component Analysis (PCA) was used to transform the data into orthogonal eigenvector components, ordered according to their contribution of variation in explaining the dataset. The first 50 components was used as features to train the supervised learning models.  

\subsection{Regression based correction methods}
The performance of the GAN correction was compared against two other popular \textit{post-hoc} correction methods: linear regression and Gaussian Process (GP) regression, which have previously been used to compensate for non-disease specific effects \citep{kostro2014correction,rao2017predictive, dukart2011age}. 

A regression model was learned to estimate the GM density for every voxel based on examples of subject-specific covariate and their corresponding GM density maps. The general linear model for the voxels is given as
\begin{equation}
\mathbf{y} = \beta_0 + \mathbf{X}\mathbf{\beta} + \mathbf{\epsilon}, \epsilon\sim\mathcal{N}(0,\sigma^2),
\end{equation}
where $\mathbf{y}$ is a $N\times v$ matrix, where the columns represent the observed GM concentrations of each voxels and the rows are the observations of each of the $N$ control subjects. $\mathbf{X} \in \mathbb{R}^{N\times2}$ is the design matrix representing the subjects' scanner characteristic, coded as $\{0,1\}$ and the intercept term. $\mathbf{\beta}\in\mathbb{R}^{2\times v}$ represents the effect strengths associated to the scanner for each voxel and the coefficient of the intercept. The regression parameters $\mathbf{\beta}$ were estimated for each voxel independently with only the control subjects to avoid the confounding of disease. The model was applied to new data, $\mathbf{x}^{(*)}$, to obtain a subject specific template, and was subtracted from the observed GM map to get a corrected image. 
\begin{equation}
\label{eqn:linear_regression_corr}
\hat{\mathbf{y}}_{OLS}^{(*)} = \mathbf{y}^{(*)} - \mathbf{x}^{(*)}\hat{\mathbf{\beta}}. 
\end{equation}
where $\hat{\mathbf{y}}_{OLS}^{(*)}$ is the corrected GM map of the original, $\mathbf{y}^{(*)}$ of the test example.

The GP regression correction method is analogous to Equation \ref{eqn:linear_regression_corr}.
\begin{equation}
\label{eqn:gp_regression_corr}
\hat{\mathbf{y}}_{GPR}^{(*)} = \mathbf{y}^{(*)} - (\mathbf{k}_{\mathbf{\theta}}^{(*)})^T\mathbf{K}^{-1}_\mathbf{\theta}\mathbf{y}. 
\end{equation}
$\hat{\mathbf{y}}_{GPR}^{(*)}$ and  $\mathbf{y}^{(*)}$ are the corrected and original images respectively. $\mathbf{K}_\mathbf{\theta}$ is the covariance kernel matrix of the training examples with the elements corresponding to the output of the kernel function $k_\mathbf{\theta}(\mathbf{x}_i,\mathbf{x}_j)$, for $i,j \in \{1,...,N\}$. The coefficients of the regression, $\mathbf{k}_{\mathbf{\theta}}^{(*)}$, is the kernel function values of the test example with all the training examples. The kernel used was similar to \cite{kostro2014correction} where the covariance between the input images $\mathbf{x}_i$ and $\mathbf{x}_j$ was
\begin{equation}
	k_{\mathbf{\theta},\sigma}(\mathbf{x}_i,\mathbf{x}_j) = \theta_1^2 \exp(-\theta_2^2(\mathbf{x}_i -\mathbf{x}_j)^2) + \theta_3^2 + \theta_4^2(\mathbf{x}_i)^T\mathbf{x}_j + \sigma^2\delta_{ij},
\end{equation}
where $\theta_k,k=\{1,...,4\}$ and $\sigma$ are scalar model hyperparameters, and $\delta_{ij}$ is the delta function; one if $i=j$ and zero, otherwise. The optimal hyperparameters were determined by maximising the likelihood function. 
\subsection{Support vector machine classification}
Each correction method in this report (GAN, GP regression, linear regression) was evaluated by the improvement of a learned supervised classifier in a range of problems such as scanner, gender and disease classification. This evaluation method was used because of the lack of ground truth; there were a limited number of subjects who were scanned across the two centers in similar conditions (\textit{n} = 11, see Experiment 4: Reconstruction), which was insufficient to fully appraise our correction methods. A popular technique for the classification of high dimensional neuroimaging data is the support vector machine (SVM). It has been used for classification of many neurological diseases such as Alzheimer's Disease \citep{magnin2009support, jongkreangkrai2016computer}, Huntington's Disease \citep{kostro2014correction} and schizophrenia \citep{winterburn2017can,davatzikos2005whole,koutsouleris2009use,zhang2014heterogeneity,kambeitz2015detecting}. SVMs learn a decision boundary based on labeled examples by maximising the margin between training examples and minimising the norm of the solution vector $\hat{\mathbf{w}}$, 
\begin{equation}
	\min_\mathbf{w} \frac{1}{n}\sum_{i=1}^n \max (0,1-y_i(\mathbf{w}\cdot \mathbf{x}_i -b)) + \lambda ||\mathbf{w}||^2 ,
    \label{eqn:svm}
\end{equation}
where the parameter $\lambda>0$ determines the tradeoff between increasing the margin-size and ensuring that $\mathbf{x}_i$ lies of the correct side of the margin.
Optimising Equation \ref{eqn:svm} can be rewritten as a constraint optimisation problem with a differentiable objective function in the following way, called the primal problem,
\begin{align}
 \nonumber &\min \frac{1}{n}\sum_{i=1}^n \zeta_i +\lambda ||\mathbf{w}||^2 \\
 &\textnormal{subject to } y_i(\mathbf{w}\cdot\mathbf{x}_i-b) \geq 1-\zeta_i 
 \textnormal{ and } \zeta_i \geq 0,\textnormal{ for all }  i.
\end{align}
The grey matter concentrations of each voxel was used as input for the classification. The primal solution, $\hat{\mathbf{w}}$, when using a linear SVM, is a linear combination of the input voxels and hence the spatial patterns of voxels that were relevant for the classification process can be visualised.

\subsection{Evaluation methods} \label{evaluation_methods}
% * <richard.morris@sydney.edu.au> 2018-04-04T04:36:52.138Z:
%
% ^.
The effectiveness of each correction technique  (linear regression, GP regression and GAN) was assessed by the classification performance of a Gaussian kernel SVM. Accuracy, precision and recall of the learned SVM was evaluated using 10-fold cross validation after each correction method was applied to the dataset, as well as a baseline of no correction. For robust evaluation, the results reported were obtained in the following manner: for a test fold, the performance measure (accuracy, precision, recall and specificity) was computed for each of the correction methods and baseline.  The difference of each measure was taken between baseline and the correction method. This was repeated for every test fold, collecting 10 sample sets for each method in each experiment. The average and standard deviation over the 10 sample sets was calculated for each method, and are the values reported. Significant differences in performance between each correction method and baseline were then compared by \textit{t}-test with Dunnett's correction to control the type-I error rate at alpha = 0.05.

  \begin{mdframed}[backgroundcolor=blue!20] 
      \begin{Large}
      \textbf{Box 1: Simulation with MNIST} \\
      \end{Large}
  The MNIST contains 50000 training examples of handwritten digits between 0 and 9. The training and test set was split in half, with one half being unaltered (Figure \ref{fig:mnist_data} top row) and the other half was change to have a black written digit against a white background, corrupted with Gaussian noise (Figure \ref{fig:mnist_data} second row). 

  \begin{center}
     \begin{minipage}{\linewidth}
     \centering
     \includegraphics[width = 1.0\textwidth]{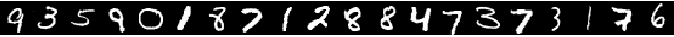}
  \includegraphics[width = 1.0\textwidth]{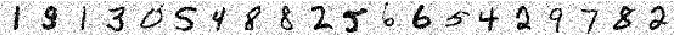}
  \includegraphics[width = 1.0\textwidth]{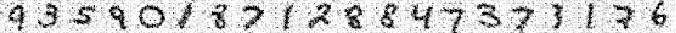}
   \includegraphics[width = 1.0\textwidth]{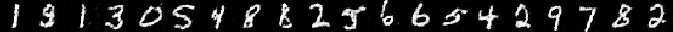}
     \captionof{figure}{\textbf{First and second row}: Sample of MNIST data set used for training. \textbf{Third and fourth row: Transformed MNIST images.}}
     \label{fig:mnist_data}
     \end{minipage}
  \end{center}
  A GAN was trained to transform the normal images to corrupted images and vice versa. The training procedure is demonstrated in Figure \ref{fig:cycle_diagram}. 
  \begin{center}
     \begin{minipage}{\linewidth}
     \centering
   \includegraphics[width = 0.9\textwidth]{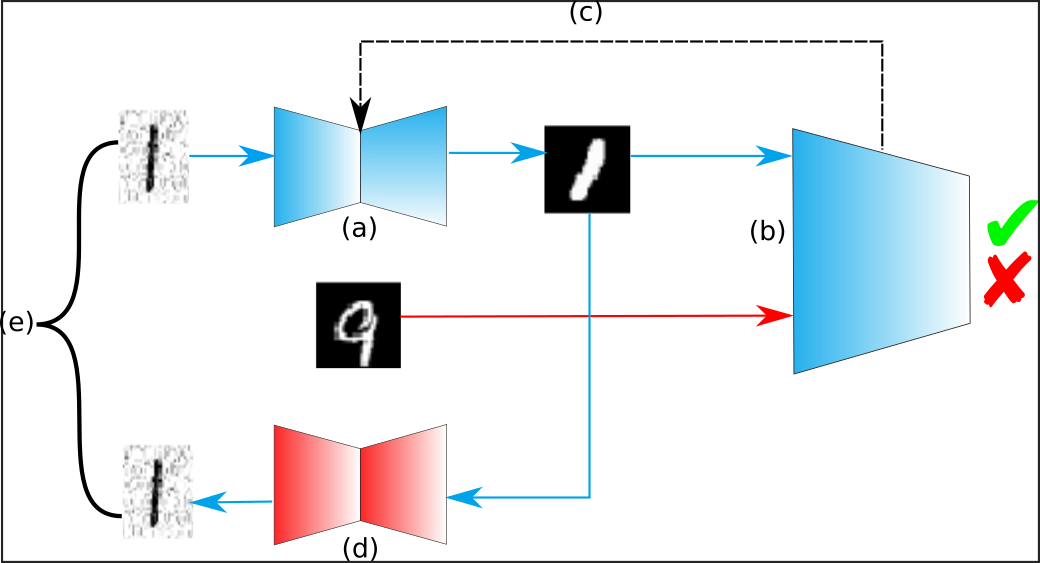}
	\captionof{figure}{}
    \label{fig:cycle_diagram}
     \end{minipage}
  \end{center}
  \textbf{(a)} A generator attempts to transform a corrupted images into a normal image. Since the generator has been initialised with random weights, in the beginning, it produces a random (noisy) image. \textbf{(b)} A discriminator attempts to classify the transformed images as fake and another image from other set as real. The digits do not necessarily have to correspond to each other.
  \textbf{(c)} The classification of the discriminator is used as information to update the generator's parameters. The discriminator, on the other hand, is told which image is fake or real and thus, is trained through supervised learning. \textbf{(d)} Another generator takes transformed image and attempts to reconstruct original image. \textbf{(e)} The original image and reconstructed image is compared and the reconstruction error is used to update both generators' parameters. \textbf{(f)} This process is mirrored for the other set of images using the same respective generators but a different discriminator. 
  
Therefore in each training cycle, the generators undergo two passes, one to transform a real image into a fake image and another to reconstruct a fake image into the original. As training progresses, the generator gradually improves its generation of images in order to fool the discriminator. At convergence, the generator is no longer able to fool the discriminator, and the discriminator is no longer able to distinguish between the observed and generated data.

  The third and fourth row of Figure \ref{fig:mnist_data} show the result of the GAN transformations on an unseen test set. These images demonstrate that the transformation still maintains the input images' most important information, its digit, and at the same time, is able to add characteristics that define the two sets of images. The GAN is able to denoise images (compare the second and bottom row of Figure \ref{fig:mnist_data}) but is also able to deterministically include features that look like Gaussian noise (compare the first row and third row).
  \label{box1:mnist}
  \end{mdframed}

\section{Results}
\subsection{Experiment 1a: Supervised classification test of scanner} \label{supervised_test}
After preprocessing, the images were converted to bias-field corrected, normalized, gray matter density maps, however site-related differences still existed in this dataset. 

\begin{figure}[!ht]
\begin{center}
 \includegraphics[width = 1\textwidth]{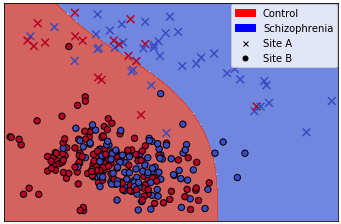}
    \end{center}
  \caption{The decision boundary, plotted in 2D, learned by a polynomial SVM when classifying diagnostic groups. The background colour represents the decision boundary. The colour of points represents the true diagnostic group membership, and the shape of points represents the scanners.}
  \label{fig:svm_decision}
\end{figure}

To illustrate the confounding influence that site-related differences can have on the ability to classify images, we initially performed a disease classification on our preprocessed (but untransformed) full dataset. Our full dataset contained images from two different groups and two different scanners. A polynomial SVM indicated the diagnostic groups were only weakly separable, and the decision boundary tended to separate scanners rather than clinical groups. Figure \ref{fig:svm_decision} shows a representation of the decision-boundary. The figure shows the decision-boundary (background colour) tends to separate shapes representing scanner differences (crosses and circles) rather than colours representing diagnostic differences (blue vs red). In particular, the crosses and circles are well-separated to the top right and bottom left of the figure, while the blue and red circles in the bottom left are intermingled. This impairs the accuracy of the model when using to predict unseen cases and favors the prediction of the sites rather than the clinical diagnosis.

\begin{figure}[!ht]
\begin{center}
 \includegraphics[width = 1\textwidth]{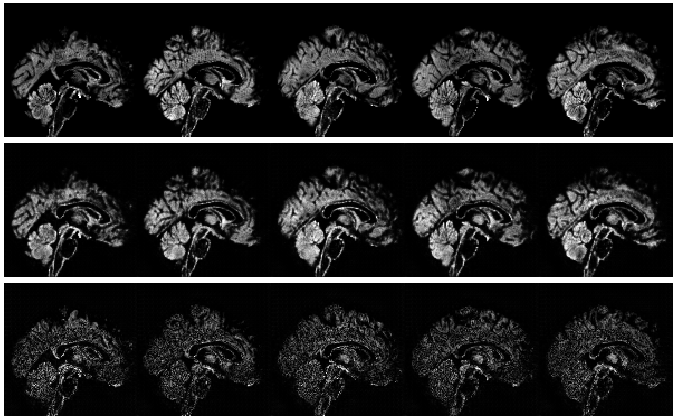}
    \end{center}
  \caption{\textbf{Top row}: Samples of images from site A. \textbf{Second row}: The result of the transformation of images from the top row using GAN. \textbf{Bottom row}: The absolute difference between the images of first and second row.}
  \label{fig:transformation}
\end{figure}

We evaluated the ability of our generative adversarial network to remove the site-related differences in our dataset. We used the mid-sagittal slice from the T1-weighted MRI of healthy subjects from site A and site B, and we merged the distribution of each image set by transforming the images from site A into images that have similar morphological characteristics as site B. 
Figure \ref{fig:transformation} shows a number of examples from the different sets and their resulting transformations. The transformed images (second row) demonstrate more consistency compared to the corresponding original images (top row). The differences between the original and transformed images, highlighted in the bottom row show significant changes in regions such as the thalamus and the brain stem.

\begin{figure}[!ht]
\begin{center}
\begin{subfigure}[t]{0.43\linewidth}
\centering
 \includegraphics[width = 1\textwidth]{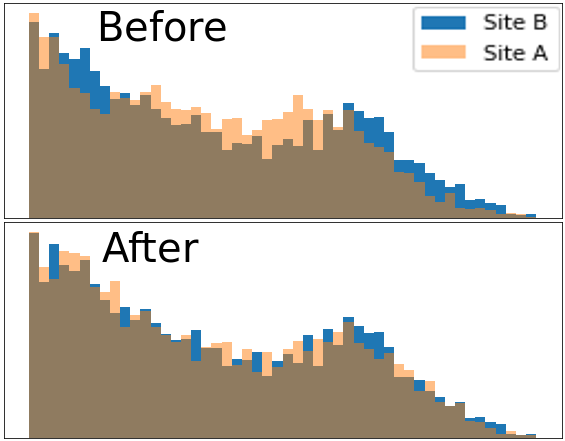}
 \caption{}
 \label{fig:pixel_distri}
 \end{subfigure}
 \hspace{0.1cm} % separation between the subfigures
\begin{subfigure}[t]{0.52\linewidth}

\centering
 \includegraphics[width = 1\textwidth,height=1.6in]{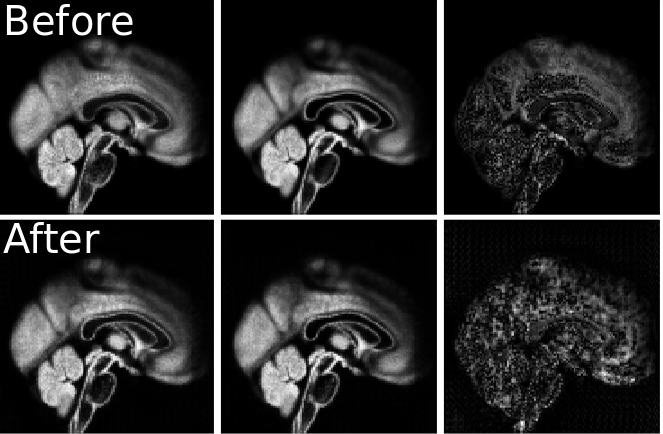}
 \caption{}.
 
 \end{subfigure}
    \end{center}
  \caption{Change in the mean image distributions of Site A and B, before (top rows) and after (bottom rows) transformation to a common distribution. \textbf{(a)} Distribution of pixel intensity before and after transformation. \textbf{(b)} Mean image from Site A (left) and Site B (middle) and the mean difference (right), before and after transformation.}
  \label{fig:mean_image}
\end{figure}

Figure \ref{fig:mean_image} demonstrates the changes in the mean image before and after the transformation using the GAN. The top rightmost image in Figure \ref{fig:mean_image} shows that the differences in the mean of Site A and B are particularly localized to the thalamus and the frontal lobe \todo{Please check if this is correct}, however after the transformation, the differences are not concentrated to a particular area of the brain. Similarly, the GAN brings the distribution of pixel intensities between Site A and B closer to each other as shown in Figure \ref{fig:pixel_distri}. 

We next conducted a supervised classification test of the dataset to determine how well the images from each site were distinguishable. A Gaussian SVM model was trained using the images from healthy controls. Table \ref{tab:classify_scanner} shows the performance of the classifier after different correctional techniques were applied to the healthy dataset, including linear regression, Gaussian regression, and our GAN transformation. The SVM was able to achieve close to 100 percent accuracy when discriminating between the two sites without any correction (99.3\% accuracy, 99.4\% precision, 99.3\% recall and 100\% specificity).  The linear correction method produced the worst outcome as the SVM was able to distinguish between the two site images with 100\% accuracy after application of this method. By contrast, the non-linear correction methods such as the GAN and GP regression reduced (but did not eliminate) the model's ability to distinguish between the sites. This suggests that the non-linear correction methods remove or minimise the site artifacts present in our dataset, with the GAN transformation producing the largest correction. 

\begin{table}[!ht]
  \centering
  \caption{Classification of scanners, using different correctional methods. Average difference in performance from baseline (no correction) across 10-fold cross-validation. Bold indicates the best performing in the category. Standard deviation in square brackets.}
\begin{tabular}{l|cccc}
    \toprule
    \textbf{Correction method} & \textbf{Accuracy} & \textbf{Precision}& \textbf{Recall} & \textbf{Specificity}\\
    \midrule
    Linear regression   &  0.007 [0.0004] & 0.006 [0.0003] & 0.007 [0.0004] & 0.000 [0.0000]\\
    GP regression    &  -0.309 [0.0243] & \textbf{-0.476} [0.0353] & -0.309 [0.0243] & -0.049 [0.0036]\\
    GAN    & \textbf{-0.386} [0.0091] & -0.389 [0.0306]& \textbf{-0.386} [0.0091] & \textbf{-0.255}[0.0151] \\
    \bottomrule
    \end{tabular}%
  \label{tab:classify_scanner}%
\end{table}

\subsection{Experiment 1b: Unsupervised classification test of scanner} \label{unsupervised_test}

\begin{figure}[!ht]
\begin{center}
 \includegraphics[width = 1\textwidth]{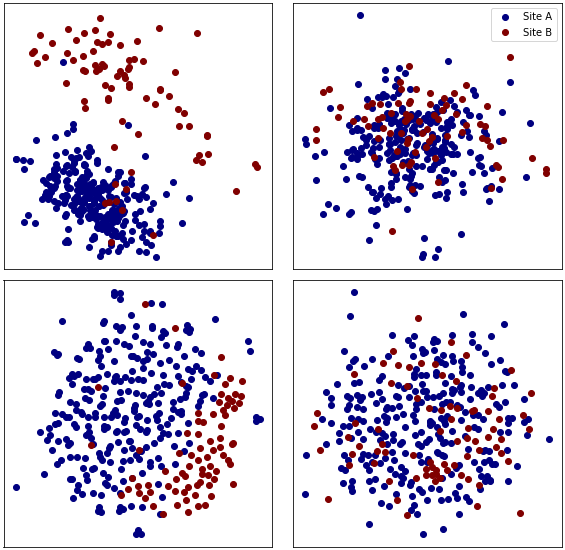}
    \end{center}
  \caption{\textbf{Left column}: Images before transformation. \textbf{Right column}: Images after GAN transformation. \textbf{Top}: PCA visualisation of the two scanner sets. \textbf{Bottom}: a t-SNE visualisation}
  \label{fig:pca_tsne}
\end{figure}

We performed unsupervised learning to determine whether any unstructured information related to site differences remained in the dataset. Figure \ref{fig:pca_tsne} shows a 2D visualisation of the differences between data sets before and after the transformation by the GAN, using two dimensionality reduction techniques: principal component analysis (PCA) and t-Distributed Stochastic Neighbor Embedding (t-SNE) \citep{maaten2008visualizing}. t-SNE, unlike PCA, is a non-linear method that is useful for exploring local neighbourhoods and finding clusters in data. If data is naively pooled (left column), there is clear separation between the datasets from each site, suggesting that these site artifacts are a possible confound and will make any interpretation of results using pooled data difficult. However, after the GAN transformation (right column), such separation has vanished and the data is akin to that generated from the same distribution.

\subsection{Experiment 2: Classification of disease} \label{class_disease}
The previous experiment demonstrated the GAN transformation method removed site-related information from our dataset on the basis of supervised and unsupervised classification methods. An important concern is whether the information loss is selective to site differences or whether other information such as that related to clinical diagnosis, is also lost. To test that, we determined whether classification of clinical diagnosis was adversely affected by any of our correction methods. A Gaussian SVM was used to classify the diagnosis of the subjects as either healthy or schizophrenia. The SVM was able to achieve over 85 percent accuracy when discriminating between clinical diagnosis without any correction (87.1\% accuracy, 89.1\% precision, 87.1\% recall and 95.7\% specificity).   Table \ref{tab:classify_disease} shows comparisons compared to baseline using each correction method (Linear and GP regression, and GAN transformation).

\begin{table}[!ht]
  \centering
  \caption{Classification of disease, using different correctional methods. Average difference in performance from baseline (no correction) over each cross validation fold is reported. Bold indicates the best performing in the category. Negative values indicate a worse result compared to baseline. Standard deviation in square brackets.}
    \begin{tabular}{l|cccc}
    \toprule
    \textbf{Correction method} & \textbf{Accuracy} & \textbf{Precision}& \textbf{Recall} & \textbf{Specificity}\\
    \midrule
    Linear regression   &  -0.003 [0.0007] & 0.000 [0.0005]  & -0.003 [0.0007]  & \textbf{0.000} [0.0010]\\
    GP regression    &  0.025 [0.0010] & 0.021 [0.0010]  & 0.026 [0.0010] & -0.042 [0.0063]\\
    GAN    & \textbf{0.037} [0.0011] & \textbf{0.028} [0.0008]& \textbf{0.038} [0.0011]  & -0.043 [0.0032]\\
    \bottomrule
    \end{tabular}%
  \label{tab:classify_disease}%
\end{table}

Linear regression was the only method to produce negative changes in accuracy, implying it non-selectively removed information from our dataset. On the other hand, GP regression and GAN transformation produced significant improvements in accuracy, with GAN producing the largest improvement in accuracy (3.7\%) when compared to base and 1.2\% compared to GP regression. The negative changes in specificity after GP and GAN correction indicate there is some improvement of classification accuracy of the schizophrenia brain images at the expense of healthy brain images.

\subsection{Experiment 3: Classification of gender}
The GAN correction appears to selectively remove information related to site differences in our dataset, without adversely affecting information related to subtle clinical differences. However anatomical differences between psychiatric groups are likely to be small, obscure and perhaps not generally representative of the morphological changes produced by our correction methods here. Furthermore, the contribution of diagnostic groups from each site in our dataset is unbalanced (e.g., see Table \ref{tab:cohort}), and there are reasonable concerns that unbalanced sampling from confounded groups may artificially inflate classification accuracy, even after weighting for unbalanced groups \citep{rao2017predictive}. To help determine the general impact of our correction methods on anatomically distinct groups, and to eliminate concerns of inflated classification accuracy due to unbalanced groups, we tested the effect of GAN correction on balanced groups. We created a dataset which balanced the group contribution from each site by randomly selecting a set of 37 male images and 37 female images from each site. Thus, we balanced both gender and site in this dataset. Male and female images from each site were then pooled together, and correction methods were applied to each dataset. We then tested whether a Gaussian SVM could classify brain images by gender. On a balanced dataset, the baseline classification accuracy of the SVM (i.e., uncorrected images) was less than 70 percent (65.2\% accuracy, 65.6\% precision, 64.5\% recall and 65.9\% specificity). The results of our correction methods are shown in Table \ref{tab:classify_gender}. The GAN corrected images improved accuracy by 15.8\% compared to baseline whereas linear regression and GP regression produced no significant difference in the classification of gender from baseline (and on average they even reduced classification performance).

\begin{table}[ht!]
  \centering
  \caption{Classification of gender, using different correctional methods. Reported values correspond to the average of the differences of each cross validation fold test between baseline (no correction) and the correction method. Bold indicates the best performing in the category. Negative values indicate a worse result compared to baseline. Standard deviation in square brackets.}
    \begin{tabular}{l|ccccc}
    \toprule
    \textbf{Correction method} & \textbf{Accuracy} & \textbf{Precision}& \textbf{Recall}& \textbf{Specificity}\\
    \midrule
    Linear regression   &  -0.015 [0.0027] & -0.018 [0.0032] & -0.016 [0.0053] & -0.014 [0.0018]\\
    GP regression    &  -0.033 [0.0026] & -0.036 [0.0022] & -0.025 [0.0056] & -0.041 [0.0071]\\
    GAN    & \textbf{0.158} [0.0332] & \textbf{0.130} [0.0362] & \textbf{0.211} [0.0310] & \textbf{0.105}[0.0576]\\
    \bottomrule
    \end{tabular}%
  \label{tab:classify_gender}%
\end{table}

\subsection{Experiment 4: Reconstruction}

\begin{figure}[!ht]
\begin{center}
 \includegraphics[width = 1\textwidth]{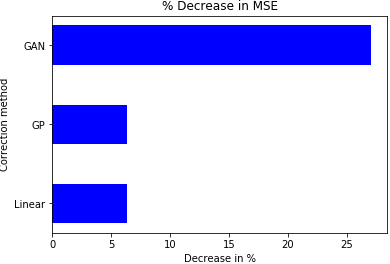}
    \end{center}
  \caption{Percentage decrease in reconstruction (MSE) error against baseline for the different correction methods.}
  \label{fig:mse}
\end{figure}

11 subjects (5 male) had undergone MRI scans at site A and site B. This allowed us to determine how similar the reconstructed images from the different methods were to images of the same brain collected at the actual site. Images from site B were corrected to site A and were compared to the actual images collected at site A for the selected subjects. The mean square error (MSE) between the corrected and actual image for each subject was calculated and was compared to baseline. Linear regression and GP regression performed similar to each other with a 6.35\% decrease in error. The GAN correction had significant improvement over the other regression methods with a 27.02\% decrease in error.

\begin{comment}
\begin{table}[ht!]
  \centering
  \caption{The percentage change from baseline in mean square error of the transformed images of site A and original site B for 11 subjects using different correction methods. A negative result indicates a worse performance compared to baseline.}
	\begin{tabular}{l|c}
    \toprule
    \textbf{Correction method} & \textbf{Change in MSE \%} \\
    \midrule
    Linear regression   &  6.34 \\
    GP regression    &  6.35 \\
    GAN    & \textbf{18.74} \\
    \bottomrule
    \end{tabular}%
  \label{tab:reconstruction}%
\end{table}
\end{comment}

\section{Discussion}
Although combining structural MRI scans from different centres provides an opportunity to increase the statistical power of brain morphometric analyses in neurological and neuropsychiatric disorders, one important confound is the potential for site differences (scanner and MRI protocol effects) to introduce systematic errors. Thus, pooling data from different sites, scanners or acquisition protocols could make the interpretation of results difficult or even decrease predictive accuracy \citep{winterburn2017can,schnack2016detecting}. These site specific differences are even more important with the growing popularity of open source data and automatic diagnostic systems using machine learning techniques. Although naively pooling data from multiple centers may increase sample size and intuitively, increase predictive accuracy, we found that the decision boundary learned by the classifier is heavily biased towards the separating hyperplane of the scanner differences rather than the true diagnostic label (See Figure \ref{fig:svm_decision}).

We proposed a novel method using deep learning to correct (unknown) site differences and experimented with data from subjects differing in clinical diagnosis or gender . The dataset was collected at two different MRI sites with different hardware and protocols. As such, our dataset probably represents larger site-related differences than previous studies which used images acquired with similar MRI protocols \citep{kostro2014correction}. Even with these large differences, we were able to remove the majority of site effects without any apparent loss in classification accuracy. These results suggest that GAN models may be a powerful method to selectively remove unwanted information from image data, without affecting the information content related to features of interest (e.g., clinical diagnosis). 

The GAN transformation left intact differences related to clinical diagnosis as well as gender. Such differences are likely to vary in magnitude relative to the site-related differences the GAN removed. For instance, VBM and MVPA indicates gray matter volume differences related to schizophrenia are small, heterogenous and widely-distributed \citep{mourao2005classifying,mourao2012individualized}. By comparison, gender differences are likely larger, with fewer major points of focus, but still widely-distributed \citep{ruigrok2014meta}. Demonstrating the selectivity of the GAN transformation against differences of varying magnitude is an important validation of the generalizability and utility of this method.

Perhaps not surprisingly, the GAN transformation produced the largest changes in the thalamus and brain stem. These regions may be more susceptible to distortions in magnetic fields and are notoriously difficult to achieve accurate image segmentation and registration during preprocessing \citep{good2001cerebral}. This is partly because it has a mix of gray and white matter which cannot be easily delineated by standard preprocessing steps. An implication of the regional variations in transformation we found is that one cannot assume that preprocessing removes all site-related differences in multi-site studies, even if bias-field correction is included. However at present it is hard to do more than speculate as to why the GAN transformation produced the changes where it did.

In comparison to other learning-based approaches, one advantage of neural networks is that no features have to be hand-crafted but instead, the model learns suitable features for the transformation during training automatically \citep{plis2014deep}. In contrast to methods such as linear regression that treat voxels independently of each other, convolutional neural networks take local information into account as they are based on image patches. The fully convolutional architecture allows for a variable number of input sizes however the quality of the generation of images may change due to the fixed receptive field of the networks.

The experiments suggest that using methods such as linear regression, and in some cases GP regression (see Table 5) are not suitable to correct for site differences. The linear regression included an intercept term to account for mean differences between sites. Yet it decreased classification accuracy when discriminating diagnostic groups and still allowed for differentiation between scanners.  On the other hand, the GAN method here was able to capture the differences between scanners, making the transformations indistinguishable between the scanner sets and improve classification accuracy compared to baseline. This suggests that site-related differences are highly nonlinear that cannot be estimated using linear methods.  

The small difference in performance between the GAN and GP regression when classifying diagnostic groups could be explained by the fact we only used a single sagittal slice from each brain in our dataset. A single slice would likely contain a relatively restricted amount of variance and hence represent a limit to the amount of information that can be learned from the data. The GAN correction, however, increased classification of gender significantly compared to GP regression. Figure \ref{fig:transformation} shows that most of the changes between original and transformed images occur around the thalamus and brain stem. Since the structural differences between gender occur in these regions \citep{ruigrok2014meta} and the result of the transformation has improved the consistency of the GM maps in those regions across scanners, this allowed the classifier to learn a decision boundary that reflected gender differences rather than variation caused by scanner differences.

\subsection{Limitations}
The major limitation of the method described here is the restriction to 2D images. That is, the current training dataset only included a small set of mid-sagittal slices rather than the entire MRI brain volume, and the test dataset only included a single mid-sagittal slice from each volume. Future work is planned to generalize this method to 3D datasets (e.g., MRI brain volumes). The extension to brain volumes could include similar techniques proposed by \cite{wu2016learning} where convolutions are performed using 3D kernels instead of 2D. However, the extension to 3D convolutional networks is not straightforward as they require more kernels than can fit on currently available hardware, and so require advanced cache management for back propagation. An alternative method is to split volumes into 2D slice data which is used to train a 2D network. Although, this loses contextual information provided by the third dimension, this is considered as a form of data augmentation and has proved very successful in tasks such as brain segmentation \citep{gonzalez2016review} However, given the massive scientific gains offered by a valid method to pool datasets in a post-hoc manner, we also hope the details we describe here will inspire other researchers to pursue the same aim, and help any  researchers currently developing a similar solution. For this reason, all data, code and models used in the present report are provided for download at  \url{https://github.com/harrisonnguyen/mri_gan}.

One advantage of conventional regression methods to correct confounds is that they allow for the inclusion of subject-specific covariates such as age and sex. The proposed GAN on the other hand, does not control for covariates and only learns a mapping between scanners while maintaining subject variation. Instead, these covariates must be included as a pre- or post- processing step using standard regression techniques. The inclusion of covariates within the GAN is left as future work.

\section*{References}

\bibliography{mybibfile}

\end{document}